\documentclass[10pt,twocolumn,letterpaper]{article}

\usepackage{cvpr}              

\definecolor{cvprblue}{rgb}{0.21,0.49,0.74}
\usepackage[pagebackref,breaklinks,colorlinks,allcolors=cvprblue]{hyperref}

\def\paperID{*****} 
\def\confName{CVPR}
\def\confYear{2025}

\title{HMVLM: Multistage Reasoning-Enhanced Vision-Language Model for Long-Tailed Driving Scenarios}

\author{
Daming Wang \quad
Yuhao Song$^*$ \quad
Zijian He \quad
Kangliang Chen \quad
Xing Pan \quad
Lu Deng \quad
Weihao Gu$^*$ \\
HAOMO.AI \\
{\tt\small songyuhao2008@gmail.com guweihao@haomo.ai}
}

\begin{document}
\maketitle
\begin{abstract}
We present \textbf{HaoMo Vision--Language Model (HMVLM)}, an end-to-end driving framework that implements the \emph{slow} branch of a cognitively inspired fast–slow architecture.  
A fast controller outputs low-level steering, throttle, and brake commands, while a slow planner-a large vision-language model-generates high-level intents such as "yield to pedestrian" or "merge after the truck" without compromising latency.  
HMVLM introduces three upgrades:
\textbf{(1) selective five-view prompting} with an embedded \(4\,\mathrm{s}\) history of ego kinematics,  
\textbf{(2) multi‑stage chain‑of‑thought (CoT) prompting} that enforces a Scene Understanding\,\(\rightarrow\)\, Driving Decision\,\(\rightarrow\)\ Trajectory Inference reasoning flow, and  
\textbf{(3) spline-based trajectory post-processing} that removes late-stage jitter and sharp turns.  
Trained on the Waymo Open Dataset, these upgrades enable HMVLM to achieve a \textbf{Rater Feedback Score(RFS) of 7.7367}, securing \textbf{2\textsuperscript{nd} place} in the 2025 Waymo Vision-based End-to-End(E2E) Driving Challenge and surpassing the public baseline by \textbf{2.77\%}.
\end{abstract}    
\section{Introduction}

\emph{Modular} pipelines inflate architectural complexity, incur cumulative information loss at every interface, and are difficult to optimise globally because each module pursues its own objective.  
In contrast, an \emph{End-to-End} (E2E) approach makes the entire chain differentiable, allowing direct optimisation for the final control task, collapsing hand-tuned interfaces, sharing one backbone for efficiency, and leveraging data-driven scaling that can unlock emergent abilities.  

The \emph{2025 Waymo Open Dataset Vision-based End-to-End Driving Challenge} curates \textbf{4,021} \(20\,\mathrm{s}\) driving segments that focus on \emph{long-tail} hazards—construction detours during marathons, pedestrians falling from scooters, debris on freeways—events that occur in fewer than \(<0.003\%\) of daily driving.  This setting provides a rigorous arena for evaluating the robustness and generalisation of E2E methods on rare yet safety-critical scenarios.

Purely fast controllers such as VAD\citep{jiang2023vad} compress the scene into vector tokens and refresh their plan with millisecond latency, but their short temporal horizon limits commonsense reasoning. At the opposite extreme, slow-only designs like EMMA\citep{hwang2024emma} recast perception, prediction, and planning as a single multimodal VLM problem, reaching state-of-the-art motion-planning accuracy at the cost of heavy compute and inference delays. The field is therefore converging on dual fast–slow pipelines: DriveVLM-Dual\citep{tian2024drivevlm} attaches a language-based planner to a real-time stack, while Senna pairs a Senna\citep{jiang2024senna} planner with an end-to-end regressor, simultaneously harvesting VLM commonsense and deterministic control. Following this trend, we also adopt a dual-system architecture. In this competition submission we expose only the slow system, a low-frequency, VLM-powered decision module responsible for semantic understanding and strategic reasoning in complex scenarios.

\textbf{HMVLM} serves as the VLM–driven slow planner in our fast–slow architecture.  
Trained solely on the Waymo Open Dataset, it attains a \textbf{Rater Feedback Score of 7.7367}, securing \textbf{2\textsuperscript{nd} place} in the 2025 Vision-based End-to-End Driving Challenge and outperforming the public baseline by \textbf{2.77\%}.  
Our work delivers three key contributions:

\begin{enumerate}[label=\arabic*)]
    \item \textbf{Selective multi-view prompting with kinematic context}. Five surround-view images, together with a \(4\,\mathrm{s}\) history of ego velocity and acceleration, are injected directly into the prompt, reducing input bandwidth while improving motion-prediction fidelity.
    \item \textbf{Multi‑stage chain‑of‑thought (CoT) prompting}. Special tokens enforce a three-stage reasoning 
    flow. Scene Understanding\,\(\rightarrow\)\, Driving Decision\,\(\rightarrow\)\ Trajectory Inference, providing interpretable intermediate text that boosts human-rater trust.
    \item \textbf{Spline-based trajectory smoothing}. Post-processing removes late-stage oscillations and sharp kinks, markedly reducing collision events.
\end{enumerate}

\begin{figure*}[!h]
    \centering
    \includegraphics[width=\textwidth]{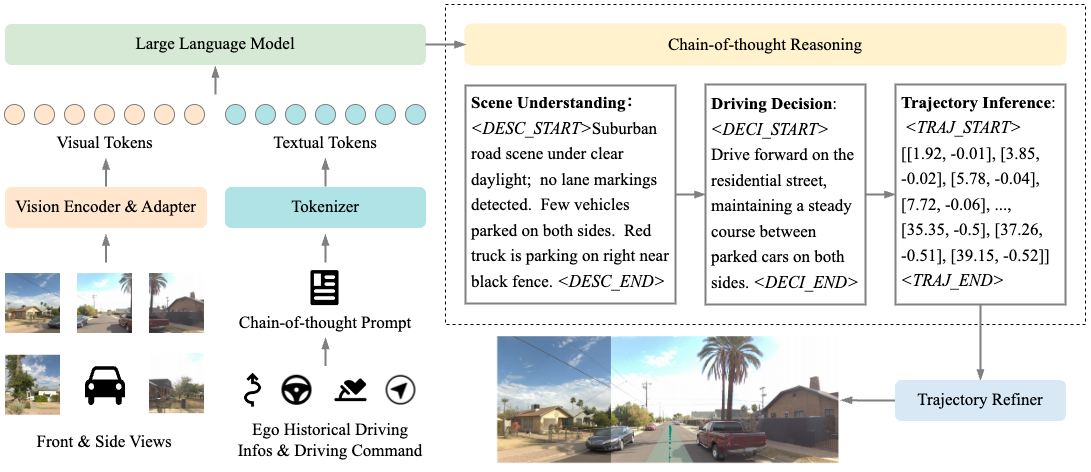}
    \caption{Overall diagram of HMVLM.}
    \label{fig:HMVLM}
\end{figure*}

\renewcommand{\thefootnote}{\fnsymbol{footnote}}
\footnotetext[1]{Corresponding authors.}
\section{Methodology}

HMVLM is an end-to-end autonomous driving framework built upon a general-purpose vision-language model. The proposed method progressively reasons from scene understanding to driving decisions and trajectory inference by leveraging multi-stage Chain-of-Thought (CoT) prompting\cite{wei2022chain} and trajectory refinement. Figure \ref{fig:HMVLM} illustrates the overall pipeline.

\subsection{Network Design and data pipeline}
HMVLM takes multi-view images as input. To balance computational efficiency and driving performance, we select three front-facing views along with left and right side views. The output is a structured, multi-stage reasoning result that includes scene understanding, driving decision-making, and trajectory prediction. For baseline model selection, we conducted a systematic comparison and evaluation of several open-source vision-language models in terms of visual perception, hallucination tendencies\cite{liu2024survey}, and reasoning capabilities under general scenarios\cite{duan2024vlmevalkit}. These dimensions are critical for autonomous driving, where safety and reliability are paramount. The overall architecture is customized and extended from the general-purpose foundation model Qwen2.5-VL-3B\citep{bai2025qwen2}.

Multimodal image-text pairs serves as the core training fuel for HMVLM. This work leverages a pre-trained large language model to perform a large-scale automated annotation on the Waymo Open Dataset, constructing two key VQA tasks aligned with the reasoning stages of HMVLM: scene understanding and driving decision making.

For scene understanding, prompt engineering is used to guide model attention toward driving-relevant elements. The model is instructed to output structured natural language descriptions that follow a specific tone and style. For the driving decision task, we incorporate ground truth trajectory data, navigation instructions, and the historical state of the ego vehicle into the model input. This helps the labeling model infer realistic human-like driving intentions and generate high-quality visual-text-action aligned annotations, which are crucial for downstream tuning and structured reasoning in HMVLM.

\subsection{Multi-stage Reasoning}
In the large model research community, it has been repeatedly demonstrated that Chain-of-Thought (CoT) reasoning significantly boosts the quality of model outputs\cite{xu2024llava}. HMVLM leverages this insight to enhance planning performance in autonomous driving. It decomposes the answer generation process into three structured reasoning stages:

\begin{itemize}

\item \textbf{Scene Understanding}. In the initial stage, HMVLM provides a global summary of the current multi-view visual input, with a particular focus on driving-relevant factors such as salient obstacles, traffic lights, lane markings, weather, and visibility.

\item \textbf{Driving Decision}. Building upon the visual understanding, the model infers driving intentions based on the vehicle’s historical states and navigation goals, expressed in natural language to describe upcoming maneuvers, such as ``prepare to turn left and slow down to yield''.

\item \textbf{Trajectory Inference}. Finally, the model translates the preceding reasoning into a structured output, a sequence of future trajectory points in the bird’s-eye view (BEV) coordinate space.

\end{itemize}

Specifically, enabling CoT-style reasoning in HMVLM requires careful design of the data flow. Instead of relying on lengthy multi-turn dialogue-style inference, HMVLM introduces dedicated special tokens to structurally define its three-stage reasoning process\cite{jiang2025alphadrive}. This not only reduces redundancy but also enhances the model’s ability to interpret and generate semantically aligned outputs.

The expected output of the model follows a structured format:
\texttt{<DESC\_START>} ``textual description of scene understanding'' \texttt{<DESC\_END>}\texttt{<DECI\_START>} ``natural language driving decision'' \texttt{<DECI\_END>}\texttt{<TRAJ\_START>} ``predicted trajectory waypoints'' \texttt{<TRAJ\_END>}.

This structured reasoning paradigm enhances both the interpretability and robustness of HMVLM’s decision-making. By explicitly disentangling visual understanding, intention, and motion generation, the model is better equipped to generalize across diverse driving scenarios.


\subsection{Trajectory Refinement}

To support HMVLM’s structured reasoning and enhance the reliability of predicted trajectories, trajectory refinement is employed to address potential kinematic discontinuities and irregularities in trajectory outputs. A smoother trajectory reduces unnecessary vehicle maneuvers and enhances safety and passenger comfort.

We implement an adaptive Savitzky-Golay filtering\cite{schafer2011savitzky} approach combined with key-point preservation. After initially removing outliers using z-score thresholding\cite{li2025developing}, the algorithm performs smoothing with adaptively selected window sizes. Critical geometric features, identified by significant directional changes exceeding 25 degrees, are explicitly preserved via weighted averaging. Additionally, trajectory endpoints are strictly maintained to ensure positional accuracy and continuity with planned routes, providing a realistic and robust trajectory prediction for autonomous driving scenarios.

\section{Experiment}
\subsection{Implementation Details}
HMVLM is built on the Qwen2.5‑VL‑3B model, which comprises the Qwen2.5 language model, a native dynamic-resolution Vision Transformer (ViT) vision encoder, and an MLP-based vision-language merger. The model is fine-tuned for 3000 iterations on the training and validation splits of the Waymo End-to-End Driving Dataset, using 8 A100 GPUs. Training is conducted with full fine-tuning using DeepSpeed ZeRO-3~\cite{rasley2020deepspeed}, employing a learning rate of 2.0e-5 with a cosine scheduler, the AdamW optimizer, a batch size of 8, and a weight decay of 0.01. Special tokens indicating the start and end of scene understanding, driving decisions, and trajectory inference are introduced alongside built-in tokens from Qwen2.5-VL. The input consists of images captured from front, front-left, front-right, side-left, and side-right camera views.

For model inference, we utilize VLLM~\cite{kwon2023efficient} as the backend with a temperature setting of 0.01, top-p sampling at 0.7, and top-k sampling at 50. Due to the inherent stochasticity of trajectory predictions from vision-language models (VLM), inferred trajectories occasionally do not match the required length of 20 points precisely. To mitigate this issue, we apply a straightforward yet effective post-processing step, trimming excessive predictions and completing shorter trajectories based on a constant-velocity assumption. These automatically completed trajectories are subsequently refined using a dedicated trajectory refiner, further enhancing their prediction quality.

\subsection{Main results}


HMVLM consistently maintains a leading position across the majority of RFS, demonstrating robust performance in diverse driving scenarios. Although its ADE scores are slightly higher compared to some competitors, HMVLM achieves second place overall in the critical RFS metric. This indicates that our VLM-based planner, fine-tuned exclusively through imitation learning, effectively generalizes to challenging and uncommon situations, precisely the capability that the RFS metric aims to capture beyond the narrower focus of ADE.

\begin{table}[!h]
\centering
\caption{Waymo E2E Driving Challenge Metrics (ADE and RFS)}
\label{tab:waymo_metrics}
\begin{tabular}{l|c}
\toprule
\textbf{Metric Category} & \textbf{Value} \\

\midrule
\multicolumn{2}{l}{\textbf{RFS}} \\
\cmidrule(lr){1-2}
\quad \textbf{Overall} & \textbf{7.7367} \\
\quad Spotlight & 6.7269 \\
\quad Construction & 8.6663 \\
\quad Intersection & 7.9043 \\
\quad Pedestrian & 7.8578 \\
\quad Cyclist  & 7.3925 \\
\quad Multi-lane Maneuvers & 7.5607 \\
\quad Single-lane Maneuvers & 8.3563 \\
\quad Cut-ins           & 7.5826 \\
\quad Foreign objects debris           & 7.8842 \\
\quad Special vehicles & 7.9710 \\
\quad Others     & 7.2013 \\
\midrule
\multicolumn{2}{l}{\textbf{ADE}} \\
\cmidrule(lr){1-2}
\quad @3s              & 1.3269 \\
\quad @5s              & 3.0715 \\
\bottomrule
\end{tabular}
\end{table}

The model's notable strength stems from effectively leveraging the advanced reasoning capabilities inherent in large language models (LLMs). Specifically, it excels in structured scenarios like Construction (8.6663) and Single-lane Maneuvers (8.3563), illustrating its ability to interpret and respond accurately to clearly defined contexts. However, it faces more substantial challenges in complex, dynamic situations such as Spotlight (6.7269) and interactions involving Cyclists (7.3925). The foundational reasoning provided by LLMs presents promising avenues for further improvement; strategic enhancement in model prompting and targeted fine-tuning could substantially elevate performance in these intricate, less predictable scenarios.




\section{Conclusion}
We introduced HaoMo Vision Language Model (HMVLM) as the deliberative (slow) planner in a fast–slow autonomous-driving stack. Leveraging selective five-view prompting, a Scene Understanding\,\(\rightarrow\)\, Driving Decision\,\(\rightarrow\)\ Trajectory Inference chain of thought, and spline smoothing, HMVLM attains a 7.74 RFS—second place in the 2025 Waymo end-to-end challenge and 2.8 \% above the public baseline. Although its vision-language inference is computationally intensive, the model’s explicit textual reasoning streamlines debugging and fosters operator trust. Future work will pair HMVLM with a millisecond-latency fast branch, broaden temporal memory, and incorporate domain-adaptive self-supervision to curb compute cost while further enhancing safety and generalisation.  
{
    \small
    \bibliographystyle{ieeenat_fullname}
    \bibliography{main}
}


\end{document}